\documentclass{article}

\usepackage{arxiv}
\usepackage{booktabs}
\usepackage{makecell}
\usepackage[authoryear,longnamesfirst]{natbib}
\usepackage{hyperref}
\usepackage{pgfplots}
\usepackage{pgfplotstable}
\usepackage{caption}
\usepackage{multirow,array,colortbl,dcolumn,stfloats,etoolbox,expl3,xparse}
\definecolor{bblue}{HTML}{4F81BD}
\definecolor{rred}{HTML}{FFB303}
\definecolor{ggreen}{HTML}{9BBB59}
\definecolor{igreen}{HTML}{579c35}
\definecolor{ppurple}{HTML}{9F4C7C}
\pgfplotsset{
   compat=1.17,
   legend entry/.initial=,
   every axis plot post/.code={%
       \pgfkeysgetvalue{/pgfplots/legend entry}\tempValue
       \ifx\tempValue\empty
           \pgfkeysalso{/pgfplots/forget plot}%
       \else
           \expandafter\addlegendentry\expandafter{\tempValue}%
       \fi
   },
}

\usepackage{subcaption}
\usepackage{wrapfig}

\def\tsc#1{\csdef{#1}{\textsc{\lowercase{#1}}\xspace}}
\tsc{WGM}
\tsc{QE}
\tsc{EP}
\tsc{PMS}
\tsc{BEC}
\tsc{DE}

\usepackage{bigfoot}

\DeclareNewFootnote{AAffil}[arabic]
\DeclareNewFootnote{ANote}[fnsymbol]

\usepackage{etoolbox}
\makeatletter
\patchcmd\maketitle{\def\@makefnmark{\rlap{\@textsuperscript{\normalfont\@thefnmark}}}}{}{}{}
\makeatother

\makeatletter
\def\thanksAAffil#1{
  \footnotemarkAAffil\protected@xdef\@thanks{\@thanks%
        \protect\footnotetextAAffil[\the \c@footnoteAAffil]{#1}}%
}
\def\thanksANote#1{%
  \footnotemarkANote%
  \protected@xdef\@thanks{\@thanks%
        \protect\footnotetextANote[\the \c@footnoteANote]{#1}}%
}
\makeatother

\author{
 Darius Feher%
 \thanksAAffil{University of Cambridge, United Kingdom}
   \And
 Abdullah Khered%
 \thanksAAffil{University of  Manchester, United Kingdom}\\
  \And
 Hao Zhang%
 \footnotemarkAAffil[2]\\
  \AND
  Riza Batista-Navarro%
  \footnotemarkAAffil[2]\\
  \And
  Viktor Schlegel%
  \footnotemarkAAffil[2] \hspace{0.03cm}$^{,}$\thanksAAffil{Imperial College London, Imperial Global Singapore} \hspace{0.03cm}$^{,}$\thanksANote{Corresponding author: \texttt{viktor.schlegel@manchester.ac.uk}}\\
}

\begin{document}
\let\WriteBookmarks\relax
\def\floatpagepagefraction{1}
\def\textpagefraction{.001}

\title{Learning to Generate and Evaluate Fact-checking Explanations with Transformers}

\maketitle
\begin{abstract}
In an era increasingly dominated by digital platforms, the spread of misinformation poses a significant challenge, highlighting the need for solutions capable of assessing information veracity. Our research contributes to the field of Explainable Artificial Antelligence (XAI) by developing transformer-based fact-checking models that contextualise and justify their decisions by generating human-accessible explanations. Importantly, we also develop models for automatic evaluation of explanations for fact-checking verdicts across different dimensions such as \texttt{(self)-contradiction}, \texttt{hallucination}, \texttt{convincingness} and \texttt{overall quality}. By introducing human-centred evaluation methods and developing specialised datasets, we emphasise the need for aligning Artificial Intelligence (AI)-generated explanations with human judgements. This approach not only advances theoretical knowledge in XAI but also holds practical implications by enhancing the transparency, reliability and users' trust in AI-driven fact-checking systems. Furthermore, the development of our metric learning models is a first step towards potentially increasing efficiency and reducing reliance on extensive manual assessment. Based on experimental results, our best performing generative model \textcolor{black}{\textsc{ROUGE-1} score of 47.77,} demonstrating superior performance in generating fact-checking explanations, particularly when provided with high-quality evidence. Additionally, the best performing metric learning model showed a moderately strong correlation with human judgements on objective dimensions \textcolor{black}{such as \texttt{(self)-contradiction} and \texttt{hallucination}, achieving a Matthews Correlation Coefficient (MCC) of around 0.7.}

\end{abstract}
\defcitealias{rocha2021impact}{Rocha et al., 2021}
\defcitealias{vidgen2021understanding}{Vidgen et al., 2021}
\defcitealias{hanselowski2018retrospective}{Hanselowski et al., 2018}
\defcitealias{thorne2018fever}{Thorne et al., 2018}
\defcitealias{nasir2021fake}{Nasir et al., 2021}
\defcitealias{harrag2022arabic}{Harrag and Djahli, 2022}

\defcitealias{beltagy2020longformer}{Beltagy et al., 2020}
\defcitealias{raffel2020exploring}{Raffel et al., 2020}
\defcitealias{he2020deberta}{He et al., 2020}

\section{Introduction}\label{sec:introduction}
Assessing the veracity of claims is a vital capability in the modern world, but it is a task that the public is often ill-equipped to do. This is evidenced, for example, by people's vulnerability to online fake news, especially with respect to topics related to public health policies~\citepalias{rocha2021impact,vidgen2021understanding}, human contribution to climate change~\citep{taddicken2023climate}  and political elections~\citep{grossman2023electoral}. Due to targeted disinformation campaigns, many users are inadvertently spreading misinformation, without critically reflecting about its sources, as the information is often presented without further context. Since experts cannot provide contextualising explanations about the validity of a claim instantaneously, there is an opportunity for the natural language processing (NLP) community to investigate automated fact verification approaches that are capable of  generating explanations.

State-of-the-art research on fact verification has mostly focussed on the capability to identify misleading claims~\citepalias{thorne2018fever}. However, for end-users, it is important to provide explanations of why exactly a claim was identified as wrong. These explanations serve both as context for the claim and as an insight into the reasoning process that led to the veracity decision. Existing fact verification approaches rely on deep learning-based models optimised on large static datasets to automatically classify whether a claim is true or false, based on retrieved supporting evidence~\citepalias{nasir2021fake, harrag2022arabic}. More formally, given a claim \texttt{C} and the evidence \texttt{E}, a fact verification model predicts a verdict \texttt{V}, consisting of a label that represents the veracity of the claim (e.g., ``True'', ``False'', ``Partially True'', ``Unverifiable'').  It is however unclear whether end-users will accept these verdicts without further context. This is further problematic, as these models have been shown to exhibit biases inferred from the datasets they were optimised upon, for example, due to reliance on the appearance of specific keywords~\citepalias{hanselowski2018retrospective}.

\begin{figure}[!t]
    \centering
    \includegraphics[width=0.6\linewidth]{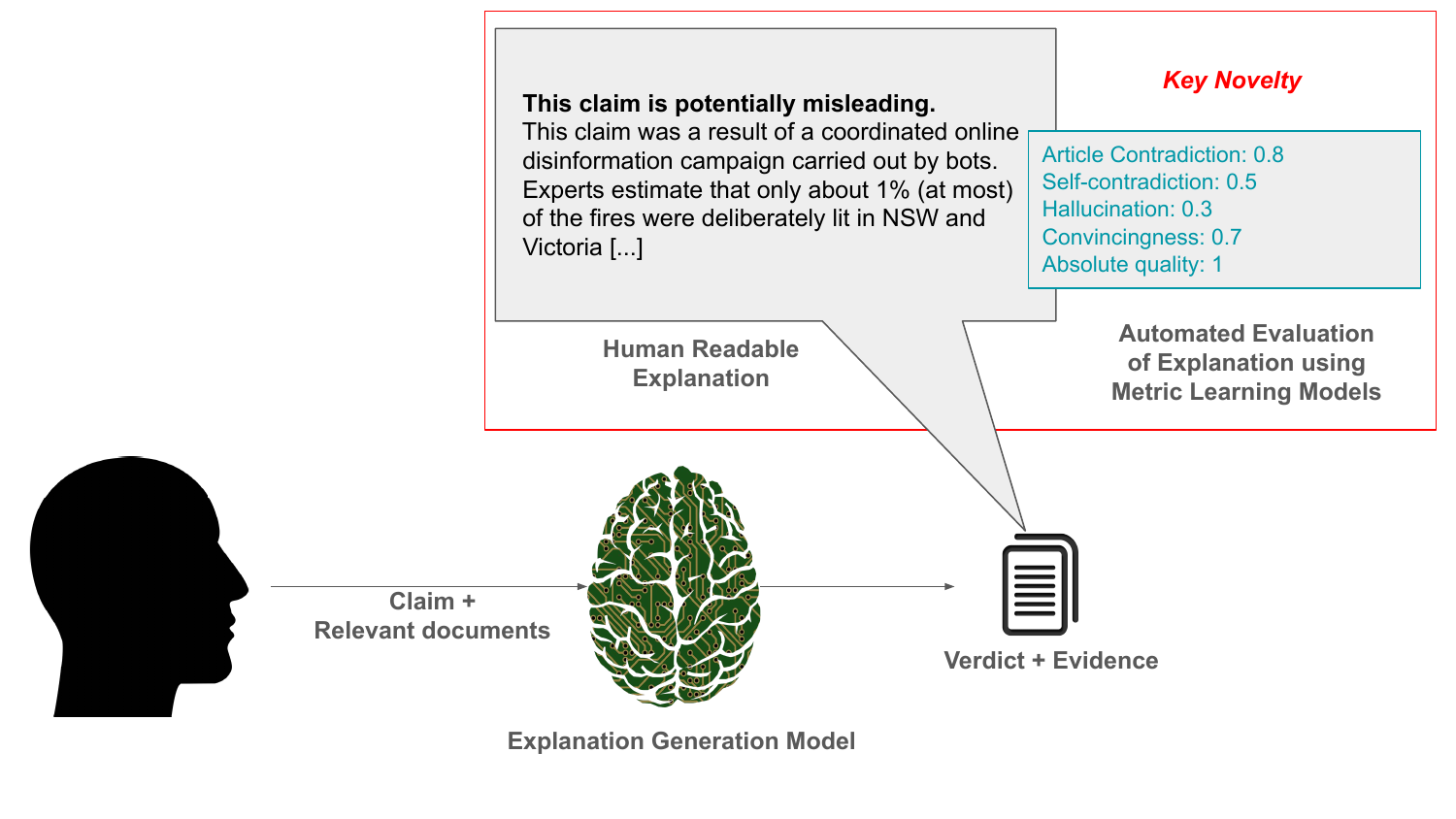}
    \caption{\color{black} Our proposed methodology outlining both explanation generation and metric learning models.}
    \label{fig:overall_architecture}
\end{figure}

\textcolor{black}{In this paper, we go beyond the state of the art in NLP-based approaches to fact verification by proposing a novel method for automated fact checking of online textual content, that contextualises and justifies its decision by generating human-accessible explanations. Importantly, we investigate the extent to which such generated explanations can be automatically evaluated, by training metric learning models on crowdsourced ratings.
}
More specifically, our research focuses on two primary tasks, depicted in Figure~\ref{fig:overall_architecture}. The first is to generate a clear, human-readable explanation that justifies the veracity of a given claim, supported by relevant evidence. The second task is to develop a method to automatically evaluate these explanations across various dimensions such as \texttt{(self)-contradiction}, \texttt{hallucination}, \texttt{convincingness} and \texttt{overall quality}, ensuring they align with human judgement standards. In line with our objectives, we seek to address the following research questions:

\textbf{RQ1:} How effectively can transformer-based models generate human-accessible explanations? This question will further be explored through an ablation study assessing the impact of dataset size and the use of imperfect evidence on a model's performance in explanation generation.

\textbf{RQ2:} To what extent can the evaluation of fact-checking explanations be automated to align with human judgements across various qualitative dimensions?

In addressing the outlined challenges, we put forward a number of research contributions, including:

\begin{enumerate}
    \item [(a)] \textcolor{black}{A novel fact-checking dataset, designed to include explanations written by journalists; this is an original gold standard dataset that we have developed by collecting claims and explanations from reliable fact-checking sources such as the BBC, Full Fact and FactCheck.}
    \item [(b)] \textcolor{black}{Transformer-based models for generating human-accessible explanations; we fine-tuned existing pretrained generative models such as the Text-to-Text Transfer Transformer \citepalias{raffel2020exploring} and Longformer Encoder-Decoder \citepalias{beltagy2020longformer} on our own fact-checking dataset, to produce new models capable of generating fact-checking explanations that bear a high level of similarity with ground truth explanations.}
    \item [(c)] \textcolor{black}{A dataset of human annotations corresponding to judgements of the quality of fact-checking explanations; this is another original dataset that we have created with the aid of crowdsourcing.}
    \item [(d)] \textcolor{black}{An automated metric learning model trained to assess explanation quality in a way that is closely aligned with human judgement standards across multiple dimensions; this model is the result of fine-tuning existing pretrained DeBERTa models \citepalias{he2020deberta} on our crowdsourced dataset of explanation quality ratings.}

\end{enumerate}

These contributions collectively push the boundaries of explainable automated fact-checking, enhancing both the generation and evaluation of explanations to meet human interpretability standards more effectively.

\defcitealias{glass2008toward}{Glass et al., 2008}
\defcitealias{krishna2021hurdles}{Krishna et al., 2021}
\defcitealias{bruhlmann2020quality}{Br{\"u}hlmann et al., 2020}
\defcitealias{schlegel2020beyond}{Schlegel et al., 2020}
\defcitealias{thorne2019evaluating}{Thorne et al., 2019}
\defcitealias{shu2019defend}{Shu et al., 2019}
\defcitealias{brown2020language}{Brown et al., 2020}
\defcitealias{fan2019eli5}{Fan et al., 2019}
\defcitealias{sellam2020bleurt}{Sellam et al., 2020}
\defcitealias{alhindi2018your}{Alhindi et al., 2018}
\defcitealias{atanasova2020generating}{Atanasova et al., 2020}
\defcitealias{fan2020generating}{Fan et al., 2020}
\defcitealias{chen2022generating}{Chen et al., 2022}
\defcitealias{dai2022ask}{Dai et al., 2022}
\defcitealias{pan2023qacheck}{Pan et al., 2023}
\defcitealias{yang2022explainable}{Yang et al., 2022}
\defcitealias{lakhotia2020fid}{Lakhotia et al., 2020}
\defcitealias{gad2019exfakt}{Gad-Elrab et al., 2019}
\defcitealias{nikopensius2023reinforcement}{Nikopensius et al., 2023 }
\defcitealias{kotonya2020explainable2}{Kotonya et al., (2020b)}
\defcitealias{yao2023end}{Yao et al., (2023)}
\defcitealias{zhang2019bertscore}{Zhang et al., 2019}
\defcitealias{alhindi2018your}{Alhindi et al., 2018}
\defcitealias{papineni2002bleu}{Papineni et al., 2002}
\defcitealias{zhu2018texygen}{Zhu et al., 2018}
\defcitealias{jelinek1977perplexity}{Jelinek et al., 1977}
\defcitealias{rony2022rome}{Rony et al., 2022}
\defcitealias{liu2023gpteval}{Liu et al., 2023}
\defcitealias{zhong2022towards}{Zhong et al., 2022}
\defcitealias{althabiti2023generative}{Althabiti et al., 2023}
\defcitealias{nauta2023anecdotal}{Nauta et al., 2023}
\defcitealias{jin2023xai}{Jin et al., 2023}
\defcitealias{ouyang2022training}{Ouyang et al., 2022}
\section{Related work}\label{sec:related_work}

In this section, we provide an overview of the relevant literature. First, we present a summary of methods employed for fact checking, followed by an outline of existing datasets for this task. Furthermore, existing methodologies used for the automated evaluation of Natural Language Generation (NLG) methods are detailed.

\subsection{Explainable fact-checking approaches}
Existing explainable approaches to fact checking can be categorised into four distinct groups. First, there are methods that provide an explanation by extracting sentences from the evidence used to check the veracity of a claim. This process can also be seen as \textit{extractive summarisation}~\citepalias{alhindi2018your, lakhotia2020fid, atanasova2020generating, fan2020generating}.
While these endeavours are scientifically justified and useful, their real-world application is potentially limited, as they typically do not provide a human-accessible way to intuitively understand and reconstruct the reasoning behind the system’s prediction. The ability to explain its logic is however a key property for developing trust in an autonomous system~\citepalias{glass2008toward}. Next, some approaches focussed on generating explanations by using \textit{question answering (QA)} as a proxy task~\citepalias{chen2022generating, dai2022ask, yang2022explainable, pan2023qacheck}. While these approaches improve explainability, they face challenges such as longer response times due to reliance on APIs and large language models (LLMs)~\citepalias{pan2023qacheck}, lack of representation due to the domain specificity of datasets, difficulty in aligning with human judgements~\citepalias{chen2022generating}, and potential error propagation from inaccurately generated questions~\citep{dai2022ask}. Additionally, the relevance of these questions can be limited, given that claims are usually short, hence, lacking context. Furthermore, another way of generating explanations is by using information available in \textit{knowledge graphs}~\citepalias{gad2019exfakt, nikopensius2023reinforcement}. In this method, the explanation consists of paths used by the agent to fact-check the claim. However, while useful, they can be complex for users to interpret. Finally, another approach that leverages advancements in NLG is the generation of explanations that are easy to understand by humans, framing the task as \textit{abstractive summarisation}, as was proposed by~\citetalias{kotonya2020explainable2} and \citetalias{yao2023end}. However, the former is using a healthcare-specific dataset for training, and both use traditional (proxy-based) NLG evaluation metrics such as \textsc{Rouge} or \textsc{BERTScore}~\citepalias{zhang2019bertscore}.



\defcitealias{kotonya2020explainable2}{Kotonya et al., 2020b}
\defcitealias{yao2023end}{Yao et al., 2023}
\subsection{Fact-checking datasets}
Most of the existing fact-checking datasets include the claim being checked, the evidence article and the veracity label, yet lack a justification or explanation of the truthfulness of the claim~\citep{thorne2018fever, hanselowski2019richly}.
\interfootnotelinepenalty=10000
Conversely, some datasets contain explanations written by journalists, but they are multimodal and thus include both text and images~\citepalias{yao2023end}, or are catering to particular domains such as healthcare~\citepalias{kotonya2020explainable2} or politics~\citepalias{alhindi2018your}.
A recently released dataset called FactEx~\citepalias{althabiti2023generative}, for instance, includes journalist explanations from \texttt{politifact.com}, a platform for fact-checking claims by politicians.

Furthermore, existing fact verification datasets may exhibit quality issues because they were gathered via crowdsourcing~\citepalias{bruhlmann2020quality, schlegel2020beyond}. Crowdsourced datasets have been shown to exhibit dataset artefacts such as arbitrary expressions that cue the ground truth label~\citepalias{thorne2019evaluating}.
As a result, models optimised on these datasets learn to exploit these cues rather than reliably performing the task.

\defcitealias{althabiti2023generative}{Althabiti et al., (2023)}
\subsection{NLG automated evaluation metrics}
In the field of NLG, automated metrics play a crucial role in evaluating the quality of generated text. These metrics can be broadly classified into two categories: task-agnostic and human-aligned. Examples of the former include perplexity~\citepalias{jelinek1977perplexity}, \textsc{Bleu}~\citepalias{papineni2002bleu}, \textsc{Rouge}~\citep{lin2004rouge}, \textsc{Meteor}~\citep{banerjee2005meteor}, or \textsc{Self-Bleu}~\citepalias{zhu2018texygen} scores, which typically measure aspects like \emph{n}-gram overlap or grammatical correctness. While they offer quick, objective and reproducible assessments, they often fail to capture the nuances of human language (e.g., coherence, relevance). Conversely, human-aligned metrics focus on how well the generated text aligns with human judgement or expectation. G-Eval~\citepalias{liu2023gpteval} utilises GPT-4 together with a chain-of-thought and form-filling framework to evaluate generated text across various dimensions including coherence, engagingness or fluency. However, there is a potential bias in G-Eval towards texts generated by LLMs, and its effectiveness depends on the availability and accessibility of these models, which incur usage costs. In contrast, other systems like UniEval~\citepalias{zhong2022towards} approach the text evaluation problem as a boolean question answering task. Moreover, another automated metric, RoMe~\citepalias{rony2022rome}, makes use of different pre-trained transformers such as ALBERT to evaluate texts based on informativeness, naturalness and quality. However, to the best of our knowledge, there are no metrics specifically optimised for qualitatively evaluating fact-checking explanations.

In our proposed research we go beyond the state of the art and address the limitations of previously reported work that we outlined above. Our main contributions are two-fold.
First, from a technical perspective, our automatic explanation generation
methods are underpinned by a model that generates human-accessible natural-language explanations: this surpasses the state-of-the-art approaches to fact verification which focus mainly on providing a verdict for a claim (i.e., true or false) and, in some cases~\citepalias{shu2019defend}, a summary extracted verbatim from relevant documents. Recent research has shown that deep learning-based models have achieved impressive performance when trained to generate free-form text conditioned on a given textual input~\citepalias{brown2020language}. Whilst these capabilities have been utilised for tasks such as machine translation, summarisation and question answering, they have been under-explored for the task of generating explanations of fact checking verdicts. This aspect of our work will thus contribute towards the emerging field of \emph{explainable fact verification}.

Secondly, previous work on generation of natural-language text such as long-form answers to questions~\citepalias{fan2019eli5} were evaluated based only on their capability to retrieve some summarised supporting information.
Although ~\citetalias{althabiti2023generative} developed models that generate fact-checking explanations based on the FactEx dataset,\footnote{We did not utilise this dataset in our own experiments as it was released only after our empirical work had been completed.} they evaluated the resulting explanations based on \textsc{Rouge} score alone, thus potentially overlooking deeper qualitative insights.
In contrast, we propose a human-centred approach to evaluating our automatically generated explanations. This will enable us to investigate the helpfulness of automatically generated explanations that contextualise fact verification results. 
To address the current challenges in evaluating natural language generation systems, we asked crowd-workers to rate the quality and convincingness of explanations generated by our system. 
Collecting a large corpus of generated explanations paired with multi-faceted human judgements of their quality allows us to learn metrics to evaluate free-text explanations rather than relying on word overlap-based metrics such as \textsc{Rouge}. This is novel, because while learned metrics for generated text exist, they do not consider quality dimensions specific to explanations~\citepalias{sellam2020bleurt}, such as their plausibility or convincingness, crucial for the user's trust and understanding of an AI system~\citepalias{ nauta2023anecdotal}.

\section{Methodology}
\begin{figure}[!ht]
    \centering
    \includegraphics[width=1.0\linewidth]{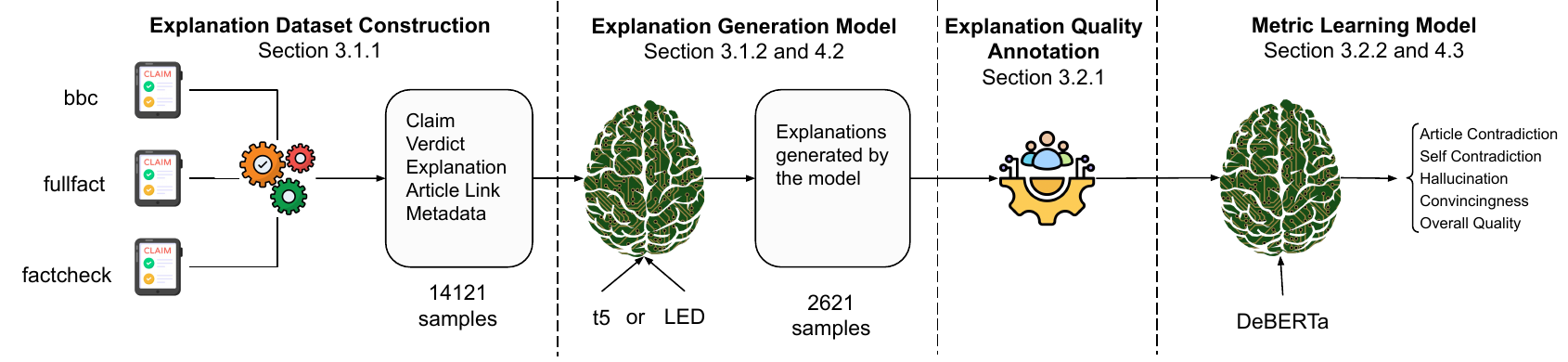}
    \caption{A visual depiction of our proposed methodology.}
    \label{fig:overall_methodology}
\end{figure}

In this section, we will discuss the approach and methods used to address the research questions outlined in Section~\ref{sec:introduction}. We first describe the data collection process, which is followed by a presentation of the methods used for the generative explanation model. We then explain how the dataset used to train the metric learning model was annotated, and outline the training methodologies employed. Figure~\ref{fig:overall_methodology} presents a visual depiction of our methodology.

\subsection{Explanation Generation}
In this section, we focus on the explanation generation component of our architecture. This includes detailing the data preparation process and the methods used for training the various explanation generation models.

\subsubsection{Data preparation and analysis}


Motivated by the limitations highlighted in Section~\ref{sec:related_work}, we created our own explanation generation dataset, with the initial step involving the collection of fact checks of various claims carried out by journalists, using Google’s FactCheck API.\footnote{\url{https://toolbox.google.com/factcheck/explorer}} Specifically, we targeted the following three fact verification outlets: \href{https://www.bbc.co.uk/}{\texttt{bbc.co.uk}},
\href{https://www.factcheck.org/}{\texttt{factcheck.org}} and
\href{https://fullfact.org/}{\texttt{fullfact.org}}.
Entries consist of \emph{(a)} the claim to be checked, \emph{(b)} the verdict of its veracity as a free-form string, \emph{(c)} the link (URL) to the article that fact-checked the claim as well as additional metadata, such as date and language. We have further enriched these details by retrieving the corresponding article using its link. For the \texttt{bbc} and \texttt{fullfact} sources, the verdicts and explanations were directly obtained from the API results (see Appendix~\ref{apd:appendix1}, Figure~\ref{fig:sub1} for an example).
Meanwhile, \texttt{factcheck} does not include any long-form explanations, hence we used the article title as the explanation (see Appendix~\ref{apd:appendix1}, Figure~\ref{fig:sub2}). It is important to acknowledge that titles may not always be precise explanations for a given claim; however, making this decision allowed us to incorporate claims published by \texttt{factcheck} into our dataset, thus significantly increasing its size. Furthermore, for the \texttt{fullfact} subset of the data, we have also collected the top ten Google Search engine hits for the claim, excluding the fullfact.org article which fact-checks the claim. This allows us to investigate the performance of a fact verification explanation generation model, when supplied with noisy evidence, i.e., Google Search engine snippets (henceforth referred to as Google snippets), which is explored in Section~\ref{sec:explanation_generation_evaluation}.




Our explanation generation dataset consists of 14,121 pairs of claim/article and explanation. We split the dataset into training and testing sets, consisting of 11,296 and 2825 examples, respectively.
To gain insight into the representation of claims verified to be true or false (and everything in between), we took the verdicts on the claims in our \texttt{factcheck} data, and mapped them to nominal categories (see Appendix~\ref{apd:appendix_mapping} for the mapping).
The resulting distribution across these nominal categories is shown in Figure~\ref{fig:label_distr}.
Notably, we excluded \texttt{bbc} and \texttt{fullfact} from this process, as they have sentence-long explanations which were not suitable for automatic mapping (see Appendix~\ref{apd:appendix1}, Figure~\ref{fig:sub1}). As expected, the verdicts are heavily skewed towards classifying claims as false. Since our aim is the evaluation of the quality of explanations of these verdicts rather than predicting the verdicts themselves, this observation is not further problematic.

\definecolor{bblue}{HTML}{4F81BD}
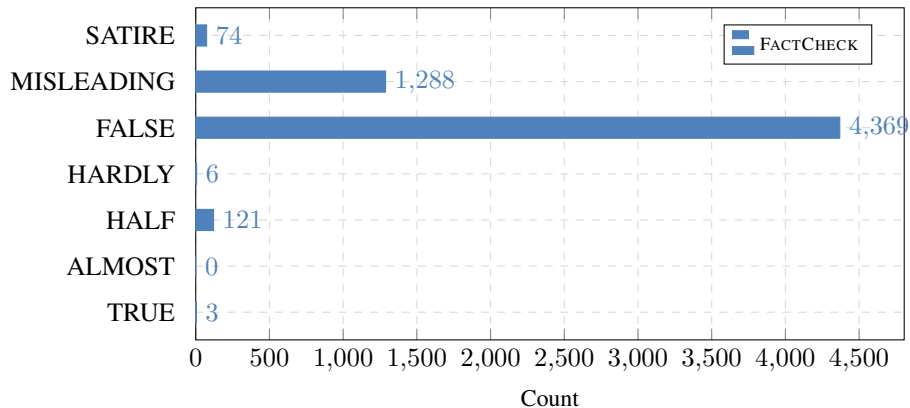
\begin{figure}[h]
    \centering
    \begin{tikzpicture}
        \begin{axis}[
            width=11cm,
            height=6cm,
            xbar, 
            xmin=0,
            xlabel={Count}, 
            symbolic y coords={TRUE, ALMOST, HALF, HARDLY, FALSE, MISLEADING, SATIRE},
            ytick=data,
            ytick style={draw=none}, 
            nodes near coords, 
            bar width=8pt, 
            ylabel style={font=\tiny}, 
            legend pos=north east,
            legend cell align={left},
            legend columns=1,
            xlabel style={font=\small},
            grid=major,
            grid style={dashed, gray!30},
            legend style={at={(0.85, 0.95)}, anchor=north, font=\small}, 
            legend cell align={left}
        ]
        \addplot[legend entry=\textsc{\scriptsize FactCheck}, color=bblue, fill=bblue] coordinates {
            (3,TRUE)
            (6,HARDLY)
            (74,SATIRE)
            (121,HALF)
            (1288,MISLEADING)
            (0,ALMOST)
            (4369,FALSE)
        };
        \end{axis}
    \end{tikzpicture}
    \caption{\color{black} Distribution of normalised labels for the \texttt{factcheck} subset.}
    \label{fig:label_distr}
\end{figure}

\begin{figure}[htb]
    \centering
    \includegraphics[width=0.66\linewidth]{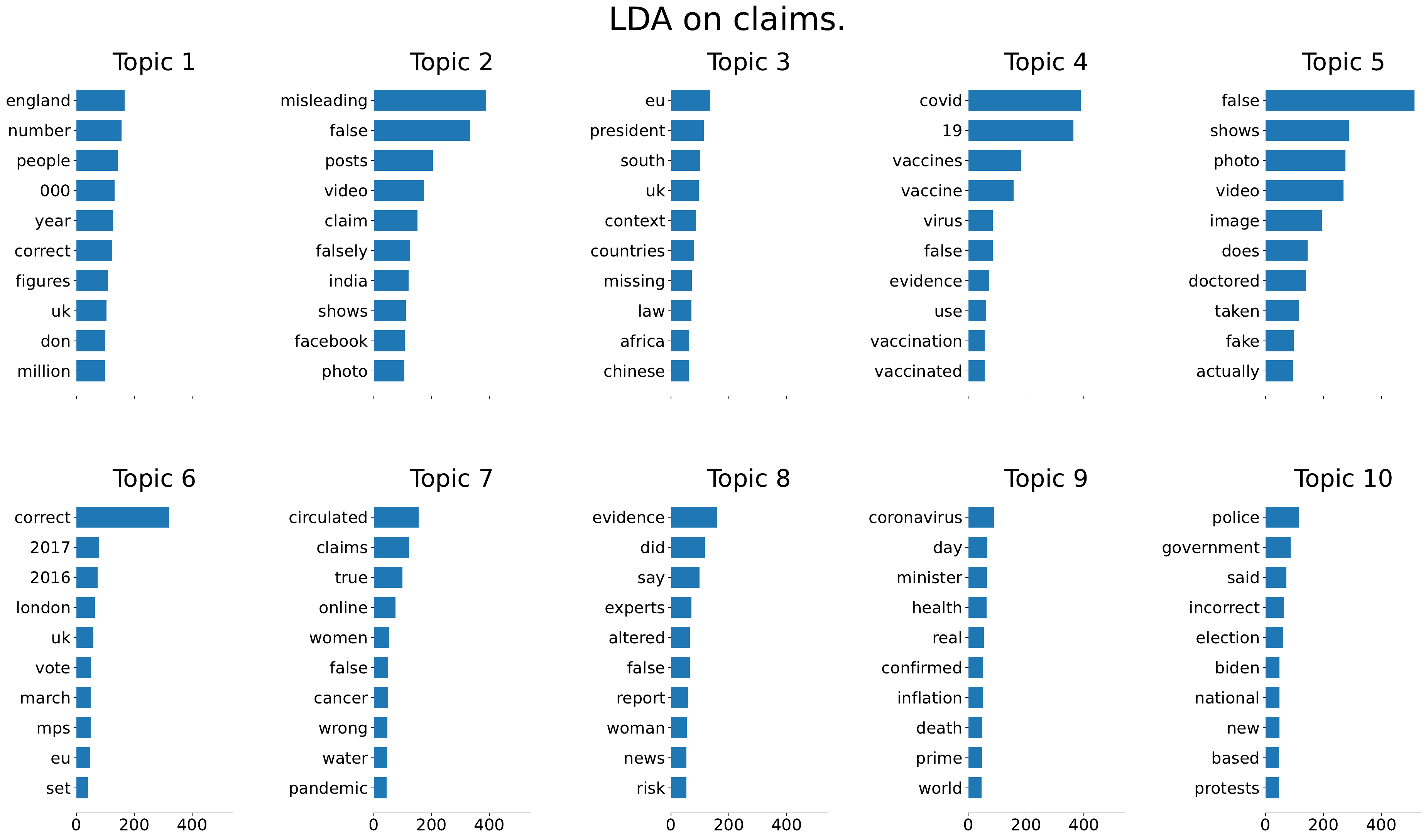}
    \caption{Topic model of all 14k claims in our dataset. Topics are described by the ten most frequent words associated with them. Note the prevalence of topical themes such as ``Covid vaccination'' or ``election and government''.}
    \label{fig:topic_distr}
\end{figure}

Moreover, Figure~\ref{fig:topic_distr} shows a topic model based on applying Latent Dirichlet Allocation (LDA) on our claims. {\color{black}More precisely, we used the LDA implementation by \texttt{scikit-learn}\footnote{\url{https://scikit-learn.org/stable/modules/generated/sklearn.decomposition.LatentDirichletAllocation.html}} with 10 topics, and the online learning method with a maximum of 10 iterations. The term-document matrix was created using a maximum of 500 features, where words appearing in more than 50\% or fewer than 10 documents were excluded. This allowed us to extract coherent themes from the data and discuss and visualise the underlying topic distribution.}

{\color{black}The results suggest that the covered topics are diverse. However, they are reflective of the media landscape of the past years, with prominent topics like ``US elections'' (Topics 1, 3, 5 and 6, with different facets therein) and ``Covid'' (Topics 4 and 8). Different modalities are covered (``photo'', ``video'' and ``shows'' and ``said'' in topics 1, 9, and 10, and 1 and 6, respectively).}

\defcitealias{williams2018broad}{Williams et al., (2018)}
\defcitealias{socher2013recursive}{Socher et al., (2013)}
\subsubsection{Methods}\label{sec:generative_model_methods}
Given a claim and some evidence, our approach is to jointly generate the veracity of the claim and provide a justification for it. Specifically, we employed a sequence-to-sequence model which takes as input a sequence \texttt{$S$}, obtained by concatenating the given claim \texttt{C} with the evidence \texttt{E} and separating them using a new line (i.e., ``\textbackslash n''). Thus, $S$ takes the following form: ``summarize: \texttt{C} \textbackslash n \texttt{E}.''  In selecting a model for our sequence-to-sequence task, we decided to initially choose \texttt{t5}, a unified Text-to-Text Transfer Transformer~\citepalias{raffel2020exploring}, which achieved strong results on tasks such as summarisation or classification. Therefore, the input sequence \texttt{$S$} is fed to \texttt{t5}, and, given its auto-regressive nature, the training objective is to maximise the log-likelihood:
$$
        \mathcal{L} = \sum_{i=1} \log(p(y_{i}|y_{1:i-1}, S; \theta))
$$

Where $y_i = $ the $i^{th}$ token in the target sequence \texttt{Y}; $y_{1:i-1} = $ all the generated tokens before the $i^{th}$ token, and $\theta = $ the parameters of the \texttt{t5} model.

Additionally, we also experimented with a Longformer Encoder-Decoder (LED) model which supports longer contexts~\citepalias{beltagy2020longformer}, i.e., up to 16K tokens.
For our preliminary experiments we decided to use the more compact version of these models, namely, \texttt{t5-base}\footnote{\url{https://huggingface.co/t5-base}} and \texttt{LED-base},\footnote{\url{https://huggingface.co/allenai/led-base-16384}} while for subsequent experiments we used \texttt{t5-large}.\footnote{\url{https://huggingface.co/t5-large}}


Furthermore, in order to test the influence the dataset size and quality on model performance, we conducted two ablation studies. Specifically, we optimised the generative models on \emph{(a)} the whole explanation dataset, \emph{(b)} the \texttt{fullfact} subset of this dataset, and \emph{(c)} the \texttt{fullfact} subset with Google snippets as evidence instead of the article, which we refer to as \texttt{fullfact-snippets}.
In each case, the claim was prepended to the evidence article or snippet, and the \texttt{t5} or \texttt{LED} encoder-decoder model was optimised to generate the ground truth verdict explanation.

To explore the potential to generate explanations from the information available on the web, we experimented with different ways of retrieving and combining search engine results as input for the sequence-to-sequence models. We compared \emph{(a)} the use of Google snippets, \emph{(b)} the use of websites that these snippets represent (\texttt{snippets-extended}), and \emph{(c)} various combinations and mapping strategies (e.g., based on exact-matching, string similarity, thresholded string similarity, different input order) for  snippets and websites. Specifically, we optimised \texttt{t5-base} and \texttt{LED-base} models on the training set of the \texttt{fullfact} portion of the dataset using different input strategies and tracked their performance on the test set both quantitatively and qualitatively. The related experiments are outlined and discussed in Section~\ref{sec:explanation_generation_evaluation}.


\subsection{Metric learning model}
In this section, we outline our approach for the metric learning model. Similar to the explanation generation model, we start by detailing the data annotation process, followed by the methods used for training the models.


\subsubsection{Data collection and annotation}\label{sec:data_collection_metric_learning}
In order to prepare our dataset for the metric learning model, we took the explanations generated by different optimised models including \texttt{t5-base}, \texttt{LED-base} and \texttt{t5-large}. The motivation behind our approach was to create a diverse dataset, reflecting varying qualities of text. Such variation is crucial for effectively training and evaluating the metric learning model, as it exposes the model to a wide range of textual qualities and complexities, enhancing its training and evaluation effectiveness. This resulted in a dataset of 2621 summaries or explanations.

In the next phase, we focussed on annotating these summaries. For this task, we engaged a group of 41 participants using the Amazon Mechanical Turk crowdsourcing platform.\footnote{\url{https://www.mturk.com/}} Participants were selected using a batched qualification task, and their compensation was determined by the quantity of annotations completed, with additional rewards granted after successful qualification. In the main task,  each of the 2,621 summaries was annotated by 3 randomly selected distinct workers. Furthermore, to reduce cognitive load, we selected the summaries based on the length of the corresponding article, with number of characters ranging between 1000---to exclude outliers and web scraping errors with lower character count---and 2500. This was to ensure that annotators were not exposed to overly long articles, which could potentially disincentivise their participation given that the compensation for this task was fixed.

The questions presented to the annotators reflected different explanation quality dimensions (see Appendix~\ref{apd:appendix_annotations_dimensions}), such as \texttt{overall quality}, which assesses the overall clarity and effectiveness of the explanation, and \texttt{convincingness}, which evaluates how persuasive the explanation is. Both these metrics are required as, for instance, an explanation could be convincing, but lack effectiveness and/or clarity. Additionally, the remaining questions aim to check for typical NLG and summarisation-related issues, such as \texttt{(self-)contradiction}---whether the explanation contradicts itself or the evidence, or \texttt{hallucination}---whether the generated text includes information or facts not present in the evidence. These quality dimensions were carefully selected based on the literature review detailed in Section~\ref{sec:related_work}. The interface used to collect the annotations is presented in Appendix~\ref{apd:appendix_annotation_interface}.

\subsubsection{Methods}\label{sec:metric_learning_model_methods}

We approached the prediction of \texttt{overall quality} as a regression task, while the prediction tasks for the other four quality dimensions were treated as binary classification tasks, training a separate model for each dimension.
Due to the presence of noise in the collected judgements (see Section~\ref{sec:metric_learning_evaluation}), we experimented with different selection strategies and trained a combination of models to predict the collected crowd judgements.
The selection strategies were based on agreement between an annotator and all their annotation peers averaged across their overall annotations, or per annotation question. Here, a score of 0.5 is the expected agreement of a random answering strategy (for binary questions). Thus, we experimented with higher thresholds such as 0.69 (i.e., on average, more than 2 out of 3 annotations are agreed upon) and 0.75 (i.e., 3 out of 4 annotations agreed upon on average). The results of these experiments are detailed in Section~\ref{sec:metric_learning_evaluation}. Additionally, to obtain training and evaluation data, we averaged the judgements of those annotators deemed eligible based on the selection strategy and split the dataset into 2100 training and 521 evaluation examples.


We experimented with transformer-based \texttt{DeBERTa-base}\footnote{\url{https://huggingface.co/microsoft/deberta-v3-base}} and \texttt{DeBERTa-xxlarge}\footnote{\url{https://huggingface.co/microsoft/deberta-v2-xxlarge}} models to investigate the extent to which model scale influences the ability to mimic human judgements. The choice of DeBERTa over BERT or RoBERTa was influenced by its superior performance on the majority of natural language understanding (NLU) tasks including natural language inference, e.g., on the MNLI dataset by~\citetalias{williams2018broad}, and binary classification, e.g., on the SST-2 dataset by~\citetalias{socher2013recursive}. The improved performance is attributed to the use of disentangled attention mechanism~\citepalias{he2020deberta}. Additionally, when employing \texttt{DeBERTa} for regression, the output layer has only one neuron, without applying an activation function, and the cross-entropy loss is replaced by the mean squared error (MSE). To obtain a more comprehensive understanding of the model's performance, we will evaluate it with both mean absolute error (MAE) and MSE metrics.
\definecolor{our_blue}{HTML}{2f5496}
\definecolor{our_green}{HTML}{38761d}
\definecolor{our_purple}{HTML}{8600e9}
\defcitealias{li2022pair}{Li et al., 2022}
\defcitealias{ji2023survey}{Ji et al., 2023}
\section{Evaluation and Results}
In this section, we present the experimental setup that was employed in this research as well as the results of evaluating the different approaches described in the previous section.

\subsection{Experimental setup}

Following the convention used in previous NLG work, especially in summarisation and machine translation, we report the performance obtained by each of the generative approaches in terms of Recall-Oriented Understudy for Gisting Evaluation (\textsc{Rouge}) metric. Specifically, we use \textsc{Rouge-1}, which measures the overlap of unigrams between the predicted and reference texts; \textsc{Rouge-2}, which evaluates bigram overlap; and \textsc{Rouge-L}, which measures the longest common subsequence between the prediction and the reference.

Furthermore, given the imbalanced nature of the classification dataset (see Section~\ref{sec:metric_learning_evaluation}), the Matthew’s Correlation Coefficient (MCC) was used to assess the performance of our \texttt{DeBERTa} binary classifiers for metric learning. MCC values range from -1 to +1, where -1 signifies complete disagreement, 0 represents a performance no better than random chance, and +1 means perfect prediction agreement.

The experiments in our study were conducted using an NVIDIA Tesla V100 GPU with 16 GB of memory, which provided the necessary computational power and memory capacity to train and test our models.

\subsection{Explanation generation}\label{sec:explanation_generation_evaluation}
As described in Section~\ref{sec:generative_model_methods}, our preliminary experiments made use of \texttt{t5-base} and \texttt{LED-base} transformer models. Thus, we firstly fine-tuned \texttt{t5-base} on the full explanation dataset and we found that the generative model is performing well on the task of contextualising the verdict, when given good evidence in the form of the full article. Table~\ref{tbl:generative_full_expl_dataset} below shows five randomly selected explanations generated by the model optimised on the full explanation, and the corresponding ground truth explanations. Generally, the model demonstrates a capability to perform claim verification and provides a sensible explanation, although omitting some details at times.

\begin{table}[h]
\caption{Examples of outputs generated by models optimised on full explanations.}\label{tbl:generative_full_expl_dataset}
\begin{tabular*}{1\textwidth}{@{}lll@{} }
\toprule
\textbf{Claim} & \textbf{Auto-generated Explanations} & \textbf{Ground Truth Explanations}\\
\midrule
\begin{tabular}[t]{@{}l@{}}330,000 people a day are being\\ tested for Covid-19 under the \\ Test and Trace programme.\end{tabular} &
\begin{tabular}[t]{@{}l@{}}This is not true. The most recent\\data shows that 366,397 people \\ were newly tested in the week that\\ ended on 22 July.\end{tabular} &
\begin{tabular}[t]{@{}l@{}}This is not true. The real number, in\\ England, is around 52,000. It will be\\slightly higher for the whole UK and if\\ antibody and surveillance tests are\\ included.\end{tabular}\\
\midrule
\begin{tabular}[t]{@{}l@{}}E. coli warning on romaine\\ lettuce in Ontario\end{tabular} &
\begin{tabular}[t]{@{}l@{}}False. Ontario article falsely warns\\ against romaine lettuce recalls\end{tabular} &
\begin{tabular}[t]{@{}l@{}}False. False article reports E. coli\\ outbreak in Ontario\end{tabular}\\
\midrule
\begin{tabular}[t]{@{}l@{}}Donald Trump tweets to\\ congratulate Ethiopian Prime\\ Minister Abiy Ahmed on Nobel\\ Peace Prize\end{tabular} &
\begin{tabular}[t]{@{}l@{}}False. No, Trump did not tweet to\\ congratulate Ethiopian Prime\\ Minister Abiy Ahmed on Nobel\\ Peace Prize\end{tabular} &
\begin{tabular}[t]{@{}l@{}}False. Trump did not tweet to\\ congratulate Ethiopian PM Abiy Ahmed \\ on his Nobel Prize\end{tabular}\\
\midrule
\begin{tabular}[t]{@{}l@{}}Video shows recent violence in\\ Orlu\end{tabular} &
\begin{tabular}[t]{@{}l@{}}False. This video shows unrest in\\ Nigeria’s southeast, not recent\\ violence\end{tabular} &
\begin{tabular}[t]{@{}l@{}}False. Video purporting to show\\ violence in Orlu is old and unrelated \\ to recent events\end{tabular}\\
\midrule
\begin{tabular}[t]{@{}l@{}}Wisconsin assembly votes to\\ withdraw its 10 electors for Joe\\ Biden\end{tabular} &
\begin{tabular}[t]{@{}l@{}}False. Wisconsin assembly did not\\ vote to withdraw its electors for\\ Biden\end{tabular} &
\begin{tabular}[t]{@{}l@{}}False. Wisconsin did not vote to\\ withdraw its US presidential electors\end{tabular}\\
\bottomrule
\end{tabular*}
\end{table}

\begin{table}[b]
\caption{Evaluation results for the \texttt{t5-base} models optimised on different datasets. Boldface indicates best results.}\label{tbl:t5_base_results}
\centering
\begin{tabular}{llll}
\toprule
\textbf{Training set} & \textsc{Rouge-1} & \textsc{Rouge-2} & \textsc{Rouge-L}\\
\midrule
\textbf{\texttt{bbc} + \texttt{fullfact} + \texttt{factcheck}} & \textbf{41.55} & \textbf{20.97} & \textbf{35.99}\\
\textbf{\texttt{fullfact}} & 35.48 & 16.01 & 28.38\\
\textbf{\texttt{fullfact-snippets}} & 25.23 & 7.39 & 21.44\\
\bottomrule
\end{tabular}
\end{table}
Next, in line with our ablation study on the impact of dataset size and quality, we fine-tuned \texttt{t5-base} on \texttt{fullfact} and \texttt{fullfact-snippets}
and the results are shown in Table~\ref{tbl:t5_base_results}. When we compare the performance of the model optimised on the full explanation dataset to that of a model fine-tuned on a smaller subset (i.e., \texttt{fullfact}), we observe a decline in performance based on the quantitative metrics. Unsurprisingly, this suggests that increasing dataset size and diversity improves the quality of produced explanations. Furthermore, when we compare the performance of the model optimised on full articles to the model optimised on Google snippets (i.e., \texttt{fullfact-snippets}), there is a significant gap, both qualitatively and quantitatively. The model optimised on snippets shows typical signs of generative models trained on noisy data, such as \textit{input copying} and \textit{self-contradictions} ~\citepalias{ji2023survey}, as shown in  Table~\ref{tbl:generative_fullfact}. Furthermore, qualitatively, the generated explanations are not as detailed as their full-article counterparts. We expect the performance to improve by introducing a stronger supervision signal (from human annotation) as well as increasing the training dataset size.


\begin{table}[t]
\caption{Examples of explanations generated by the model optimised on the \texttt{fullfact} set (Model-Full Explanations) and \texttt{fullfact-snippets} set (Model-Snippets).
Highlighted are instances of \textcolor{our_blue}{self-contradiction}, \textcolor{our_purple}{input copying} and \textcolor{our_green}{detail omission}.}\label{tbl:generative_fullfact}
\begin{tabular*}{\textwidth}{@{}llll@{} }
\toprule
\textbf{Claim} & \textbf{Model-Full Explanations} & \textbf{Model-Snippets} & \textbf{Ground Truth Explanations}\\
\midrule
\begin{tabular}[t]{@{}l@{}}Britain is the third\\ largest military power\\ in the world.\end{tabular} &
\begin{tabular}[t]{@{}l@{}}The UK has the sixth\\ largest defence budget in\\ the world, according to\\ the International\\ Institute for Strategic\\ Studies.\end{tabular} &
\begin{tabular}[t]{@{}l@{}}This is based\\ on \textcolor{our_blue}{military}\\ \textcolor{our_blue}{spending, not}\\ \textcolor{our_blue}{military}\\ \textcolor{our_blue}{spending.}\end{tabular} &

\begin{tabular}[t]{@{}l@{}}A think tank ranked the UK third in the\\ world for ``military might'' based on\\ judgements about countries’ defence\\ spending, nuclear power, infrastructure,\\ personnel and readiness. But this isn’t\\ the only way to measure military\\ strength and a number of others rank\\ the UK slightly lower.\end{tabular} \\
\midrule
\begin{tabular}[t]{@{}l@{}}Neomycin sulphate, an\\ antibiotic, is in vaccines\\ and can lead to\\ epilepsy, brain damage\\ and allergic reactions.\end{tabular} &
\begin{tabular}[t]{@{}l@{}}This isn’t in any UK\\ vaccines. It’s in many\\ vaccines, and there’s no\\ evidence it can cause any\\of these.\end{tabular} &
\begin{tabular}[t]{@{}l@{}}Neomycin\\ sulphate is in\\ some\\ vaccines,\\ but is not in all.\end{tabular} &
\begin{tabular}[t]{@{}l@{}}Trace amounts of this antibiotic may\\ end up in certain vaccines. If you are\\ allergic to it could cause an allergic\\reaction.\end{tabular}\\
\midrule
\begin{tabular}[t]{@{}l@{}}Only 138,000\\ immigrants spoke no\\ English at the last\\ census, less than 0.3\%\\ of population.\end{tabular} &
\begin{tabular}[t]{@{}l@{}}Correct. 118,000 people\\ born outside the UK\\ spoke no English or\\ Welsh.\end{tabular} &
\begin{tabular}[t]{@{}l@{}}\textcolor{our_green}{Correct.}\end{tabular}&
\begin{tabular}[t]{@{}l@{}}
Not quite correct. 138,000 people in the\\UK spoke no English at the last census,\\ about 0.3\% of the population. 118,000\\ of those people were born outside the\\ UK, 20,000 inside.\end{tabular}  \\
\midrule
\begin{tabular}[t]{@{}l@{}}Lockdowns will end\\ when healthy people\\ stop getting tested.\end{tabular} &
\begin{tabular}[t]{@{}l@{}}There is no fixed figure\\ for the asymptomatic\\ rate of people infected\\ with Covid-19, and it's\\ hard to know the precise\\ false positive rate for\\ Covid-19 tests, but all\\ the evidence suggests\\ that the number is\\ extremely low.\end{tabular} &
\begin{tabular}[t]{@{}l@{}}This is not\\ true.\\ \textcolor{our_purple}{Lockdowns}\\ \textcolor{our_purple}{will end when}\\ \textcolor{our_purple}{healthy}\\ \textcolor{our_purple}{people stop}\\ \textcolor{our_purple}{getting}\\ \textcolor{our_purple}{tested.}\\\end{tabular}&
\begin{tabular}[t]{@{}l@{}}People not getting tested will not stop\\ lockdowns from happening. The number\\ of tests, positive cases and things like\\ test positivity are all used to measure\\ the outbreak. A lack of testing might\\ lead to an extended pandemic.\end{tabular}\\
\midrule
\bottomrule
\end{tabular*}
\end{table}

Following this, the conducted experiments, aimed at investigating a model's capability to generate explanations only from information available online, indicate that expanding articles with retrieved snippets did not substantially improve performance, compared to using snippets as input alone. This is reflected in Table~\ref{tbl:rouge_scores_difference_snippet_str}, where the first row, representing the model optimised on \texttt{fullfact-snippets} or \texttt{Claim} + \texttt{Snippets}, achieves similar results to those in subsequent rows. Additionally, a qualitative analysis of the generated predictions reveals that while the \textsc{Rouge}-scores are similarly low, the semantic content of the predictions generated by models trained on datasets with different strategies can be contradictory (e.g., for the same claim, one model would predict ``False. \ldots'' while another would predict ``True. \ldots''). This result shows that relying solely on \textsc{Rouge} scores for NLG models evaluation is not sufficient, motivating the need for more qualitative metrics.
\begin{table}[t]
\caption{Evaluation results for the \texttt{t5-base} models optimised on claims from the \texttt{fullfact} training set and different strategies for utilising Google snippets. \texttt{Claim+Snippet} is the original dataset, \texttt{ExpandedEM} means search result snippets were matched with paragraphs from the linked websites only if an exact match (EM) was found, and \texttt{ExpandedLSX} means claims were matched if there was a lexical similarity (LS) of at least X between snippet and passage. Boldface indicates best results.}\label{tbl:rouge_scores_difference_snippet_str}
\centering
\begin{tabular}{@{}llll@{} }
\toprule
\textbf{Strategy} & \textsc{Rouge-1} &\textsc{Rouge-2} & \textsc{Rouge-L}\\
\midrule
Claims+Snippet & 25.23 & 7.39 & \textbf{21.44}\\
Claims+ExpandedEMOnly & \textbf{25.64} & 7.53 & 20.03\\
Claims+ExpandedLS0.3 & 25.24 & 7.37 & 19.58\\
Claims+ExpandedLS0.5 & 25.25 & \textbf{7.64} & 19.81\\
Claims+ExpandedLS0.7 & 25.04 & 7.28 & 19.59\\
\bottomrule
\end{tabular}
\end{table}

Moreover, we consider these findings as evidence that the information contained in the snippets might be insufficient, despite exploring various snippet expansion strategies. This seemed to have skewed the quality of the generated explanations, to be annotated by the crowd-workers, towards the lower-end spectrum. Thus, we decided to use explanations generated by models optimised on the full explanation dataset as input for the main annotation task. To introduce variability, we instead vary model size (e.g., \texttt{t5-large} compared to \texttt{t5-base}), architecture, and input length (i.e., \texttt{t5-base} with a limit of 1024 tokens vs the better performing \texttt{LED-base} with a limit of 2048 tokens). From Table~\ref{tbl:led_and_t5_rouge}, we can see that the \texttt{LED-base} model achieved superior performance compared to \texttt{t5-base}. We attribute this success to the model's capability to access a broader context without the need for truncation. However, \texttt{t5-large} outperforms \texttt{LED-base}, which is likely due to its significantly larger size, enabling it to capture more complex patterns. \textcolor{black}{Statistically significant differences are observed between the \texttt{LED-base} and \texttt{t5-large} models across all metrics---\textsc{ROUGE-1}, \textsc{ROUGE-2} and \textsc{ROUGE-L}--- with corresponding p-values of 0.002, 0.02, and 0.006 respectively, as determined by the paired t-test that we conducted.}

\begin{table}[t]
\centering
\caption{Evaluation results for \texttt{t5-base} and \texttt{LED-base} on the full explanation dataset. Boldface indicates best results.}\label{tbl:t5_led_base_results}
\begin{tabular}{@{}lllll@{} }
\toprule
\textbf{Model} & \textbf{Input length} & \textsc{Rouge-1} & \textsc{Rouge-2} & \textsc{Rouge-L}\\
\midrule
\texttt{t5-base} & 1024 & 41.55 & 20.97 & 35.99\\
\texttt{t5-large} & 1024 & \textbf{47.77} & \textbf{27.01} & \textbf{42.08}\\

    \texttt{LED-base} & 2048 & 46.45 & 26.01 & 40.91\\
\bottomrule
\end{tabular}\label{tbl:led_and_t5_rouge}
\end{table}

\subsection{Metric learning model}\label{sec:metric_learning_evaluation}

Considering the results obtained in the previous section (see Table~\ref{tbl:led_and_t5_rouge}) and the methodology for our metric learning, outlined in Section~\ref{sec:metric_learning_model_methods}, we opted to use the explanations generated by different models, optimised on the full explanation dataset, as input for annotation. We performed this task by following the procedure outlined in Section~\ref{sec:data_collection_metric_learning}.

\begin{figure}[!t]
    \centering
    \includegraphics[width=0.7\linewidth]{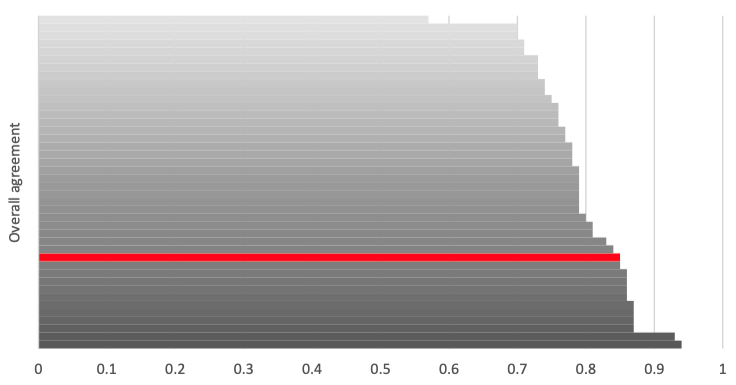}
    \caption{Average agreement of workers on all tasks. Highlighted in red is the average agreement of ChatGPT.}
    \label{fig:avg_agreement}
\end{figure}

The overall agreement statistics for the crowdsourced annotations of the generated explanations are reported in Table~\ref{tbl:annotations_results}. We note that the perfect agreement scores were low, which hints at the subjectiveness of the task (e.g., evaluating convincingness) as well as the presence of noise in the annotations. Investigating the average agreement per annotator, we find that some annotators performed only marginally better than the random selection strategy (see Figure~\ref{fig:avg_agreement}). Additionally, the agreement scores tend to be even lower if averaging per question.

\begin{table}[b]
\caption{Annotation agreement results. Note that we did not calculate the usual metrics such as Krippendorf alpha, as each item was annotated by a different set of crowdworkers. Instead, we report simple accuracy across all annotators as agreement.}\label{tbl:annotations_results}
\centering
\begin{tabular}{@{}lll@{} }
\toprule
\makecell{Category}  & \makecell{\% perfect agreement} & \makecell{\% partial agreement}\\
\midrule
        \makecell{Article Contradiction} & \makecell{0.20} & \makecell{0.72}\\
        \makecell{Self Contradiction} & \makecell{0.24} & \makecell{0.74} \\
        \makecell{Hallucination} & \makecell{0.12} & \makecell{0.66}\\
        \makecell{Convincingness} & \makecell{0.17} & \makecell{0.64} \\
\bottomrule
\end{tabular}
\end{table}

Qualitatively investigating the annotated data, we found examples of noisy annotations (e.g., annotators indicating that the explanation does contradict itself while being convincing at the same time). To reduce this noise, we regarded only annotations of those annotators whose average agreement was higher than 0.75. In some cases, this led us to situations without agreement on the binary questions (where one annotator was excluded, and the remaining two did not reach a consensus). In alleviating this issue, we took inspiration from the recent finding that large language models (e.g., GPT3 and ChatGPT) can perform annotation tasks at a comparable performance with lay annotators~\citep{kalyan2023survey}, and used OpenAI's ChatGPT-3.5-turbo API to perform a tie break on the objective questions aimed at NLG quality (i.e., contradiction and hallucination). Based on Figure~\ref{fig:avg_agreement}, it is evident that, on average, ChatGPT's agreement surpasses that of two-thirds of the crowd-workers. It is important to note that, for subjective questions (i.e., convincingness and overall quality rating), we refrained from relying on ChatGPT annotations for metric optimisation.

\begin{table}[h]
\caption{Label distribution in the binary datasets after disregarding annotations by crowdworkers with low agreement and breaking ties by ChatGPT for the first three categories, which are objective.}\label{tbl:annotation_label_distr}
\centering
\begin{tabular}{@{}lll@{} }
\toprule
\makecell{Category}  & \makecell{\# True} & \makecell{\# False}\\
\midrule
        \makecell{Article Contradiction} & \makecell{384} & \makecell{2104}\\
        \makecell{Self Contradiction} & \makecell{135} & \makecell{2486} \\
        \makecell{Hallucination} & \makecell{167} & \makecell{2454}\\
        \makecell{Convincingness} & \makecell{863} & \makecell{1343} \\
\bottomrule
\end{tabular}
\end{table}

The overall distributions of the questions after this manipulation can be seen in Table~\ref{tbl:annotation_label_distr}. The classification dataset is heavily skewed towards the label ``False,'' perhaps due to the use of strong generative baselines and comparatively short inputs---on which generative models tend to perform better. Due to the imbalanced nature of our dataset, accuracy is not a suitable metric. As a result, we evaluated the classifiers based on Matthew’s Correlation Coefficient (MCC), which takes class imbalance into account.

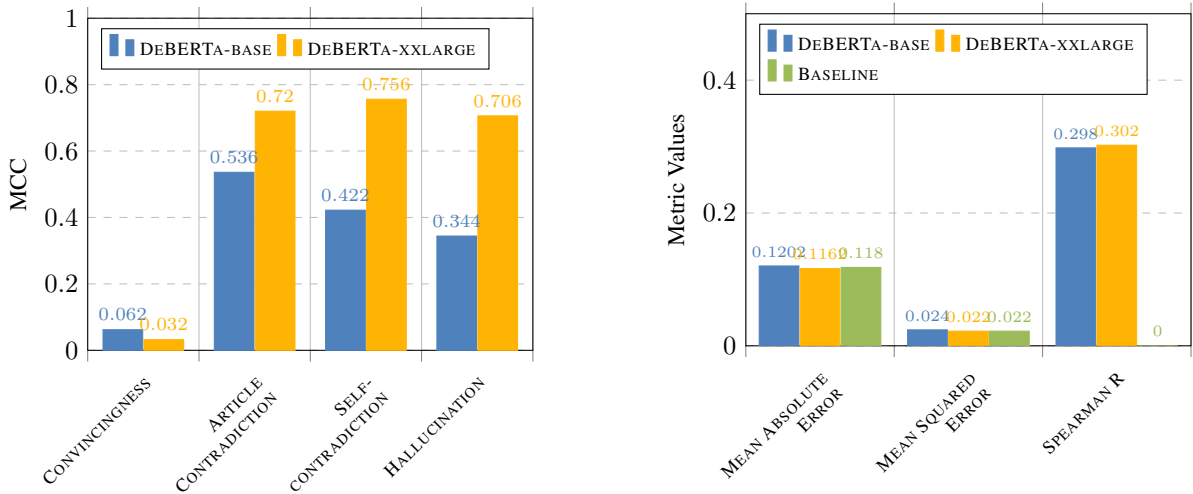
\begin{figure*}[t]
    \centering
    \begin{subfigure}[t]{0.47\textwidth}
        \begin{tikzpicture}
            \begin{axis}[
                ybar=0.5pt,
                bar width=15pt,
                width=7.5cm,
                height=6cm,
                xtick={0,1,2,3,4},
                xticklabels={Convincingness, Article\\Contradiction, Self-\\contradiction, Hallucination},
                legend pos=north west,
                legend cell align={left},
                legend columns=3,
                xmin=0, xmax=4,
                ymin=0.0, ymax=1,
                ylabel={MCC},
                ylabel style={font=\small, color=black},
                xlabel style={font=\large, color=black},
                ymajorgrids=true,
                y grid style=dashed,
                major y tick style = transparent,
                xmajorgrids=true,
                xminorgrids=true,
                x tick label style={rotate=45, font=\scriptsize\scshape, anchor=north east, align=right},
                x tick label as interval,
                nodes near coords,
                every node near coord/.append style={font=\scriptsize},
                every node near coord/.append style={/pgf/number format/.cd, fixed, precision=4}, 
            ]
            \addplot[legend entry=\textsc{\scriptsize DeBERTa-base}, color=bblue, fill=bblue] coordinates {(0.5, 0.062) (1.5, 0.536) (2.5, 0.422) (3.5, 0.344)};
            \addplot[legend entry=\textsc{\scriptsize DeBERTa-xxlarge}, color=rred, fill=rred] coordinates {(0.5, 0.032) (1.5, 0.720) (2.5, 0.756) (3.5, 0.706)};

            \end{axis}
        \end{tikzpicture}
        \caption{\color{black} Correlation of predictions by optimised models with human ratings. The score being tracked is MCC, and the majority baseline score is 0.}
        \label{fig:mcc}
    \end{subfigure}
    \hfill
    \begin{subfigure}[t]{0.47\textwidth}
        \begin{tikzpicture}
            \begin{axis}[
                ybar=0.5pt,
                bar width=15pt,
                width=7.5cm,
                height=6cm,
                xtick={0,1,2,3},
                xticklabels={Mean Absolute\\Error, Mean Squared \\Error, Spearman R},
                legend pos=north west,
                legend cell align={left},
                legend columns=2,
                xmin=0, xmax=3,
                ymin=0.0, ymax=0.5,
                ylabel={Metric Values},
                ylabel style={font=\small, color=black},
                xlabel style={font=\large, color=black},
                ymajorgrids=true,
                y grid style=dashed,
                major y tick style = transparent,
                xmajorgrids=true,
                xminorgrids=true,
                x tick label style={rotate=45, font=\scriptsize\scshape, anchor=north east, align=right},
                x tick label as interval,
                nodes near coords,
                every node near coord/.append style={font=\tiny},
                every node near coord/.append style={/pgf/number format/.cd, fixed, precision=4}, 
            ]
            \addplot[legend entry=\textsc{\scriptsize DeBERTa-base}, color=bblue, fill=bblue] coordinates {(0.5, 0.1202) (1.5, 0.024) (2.5, 0.298)};
            \addplot[legend entry=\textsc{\scriptsize DeBERTa-xxlarge}, color=rred, fill=rred] coordinates {(0.5, 0.1162) (1.5, 0.022) (2.5, 0.302)};
            \addplot[legend entry=\textsc{\scriptsize Baseline}, color=ggreen, fill=ggreen]coordinates {(0.5, 0.118) (1.5, 0.022) (2.5, 0)};

            \end{axis}
        \end{tikzpicture}
        \caption{\color{black} Correlation of predictions with human ratings for \textit{overall quality}, filtered to include only annotations with an agreement score of at least 0.75. The baseline score is determined by averaging the high-agreement annotations in the dataset\footnotemark.}
        \label{fig:mae}
    \end{subfigure}
\caption{Metric learning results.}
\label{fig:results}
\end{figure*}

We next fine-tuned \texttt{DeBERTa-base} and \texttt{DeBERTa-xxlarge} transformers on our 2100 training samples. The aim was to accurately predict \texttt{overall quality} (i.e., the first dimension), and perform binary classification across the other four quality dimensions annotated: \texttt{article contradiction}, \texttt{self-contradiction}, \texttt{hallucination} and \texttt{convincingness}. Figures~\ref{fig:mcc}~and~\ref{fig:mae} present the main findings of the best models optimised in these settings. Unsurprisingly, model scale seems to lead to consistent improvements {\color{black}(with the exception of the low-scoring subjective \texttt{convincingness} category)}. Furthermore, the improvements over the baselines are most noticeable in more objective categories such as detecting contradictions, even if the datasets are heavily imbalanced. Identifying contradictions between the summary and the main article seems to be the task where both models perform best {\color{black} on average}. This is consistent with the literature on Natural Language Inference, where models often perform on par with humans on the task of detecting whether a pair of sentences contradict each other~\citepalias{li2022pair}. \textcolor{black}{Moreover, we observe statistically significant differences ($p < 0.05$) between the \texttt{DeBERTa-base} and \texttt{DeBERTa-xxlarge} as well as between the models and the majority baseline in terms of MCC, MAE, and MSE. The p-values were calculated by re-running each experiment five times with a different train/test split, then conducting the t-test for one sample for comparison against the baseline (i.e. known population mean of 0) and a t-test for two related samples for comparison between the obtained model performance scores.
Details on hyper-parameter settings and obtained p-values are found in  Appendix~\ref{appendix:statistical_significance_metric_learning}.}

Overall, our results indicate that the prediction of human judgements remains a hard task that warrants further academic investigation, even as the best of our optimised models reach \textcolor{black}{around 0.7} MCC on more objective questions, which represents strong correlation. More subjective dimensions, such as the convincingness and overall quality of explanations are even harder, with our best models consistently outperforming statistical baselines, albeit by a small margin. {\color{black} While the gains in absolute or squared error for the overall quality prediction regression task do not yield any statistically significant improvements over the baseline, looking at the (Spearman) correlation between predictions and ground truth ratings paints a slightly different picture: both models' predictions correlate mildly with ground-truth annotations. This means that while individual errors in prediction might be high, contributing to a relatively high absolute and squared errors, their impact on the ability to correctly rank explanations from worst to best (which is what the Spearman correlation metric measures) is less pronounced.}

\footnotetext{Note that in this way, the Spearman metric is technically not defined as the predictions on the test set are constant. However adding an infinitesimally small amount of noise to each prediction will result in a metric close to zero.}



\defcitealias{sun2023query}{Sun et al., 2023}
\defcitealias{li-etal-2024-side}{Li et al., (2024)}
\defcitealias{rafailov2024direct}{Rafailov et al., 2024}
\defcitealias{roit2020controlled}{Roit et al., 2020}
\defcitealias{barai2024crowdsourcing}{Barai et al., 2024}
\defcitealias{ding2023gpt}{Ding et al., 2023}
\defcitealias{beck2023quality}{Beck, 2023}
\defcitealias{frenda2024perspectivist}{Frenda et al., 2024}
\defcitealias{fleisig-etal-2024-perspectivist}{Fleisig et al., 2024}
\subsection{Discussion}
{\color{black}
The findings of this study demonstrate significant advancements in the area of fact-checking explanation generation and evaluation. Firstly, our results indicate that it is feasible to train models to generate claim explanations effectively, which receive positive evaluations from human annotators regarding their quality. Secondly, we have shown that it is possible to develop models capable of directly predicting the quality of these generated explanations. Notably, LLMs, such as ChatGPT, exhibit a considerable degree of success in this task in a zero-shot setting, as evidenced by the high overall agreement of ChatGPT with human crowd-workers presented in Figure~\ref{fig:avg_agreement} and discussed in Section~\ref{sec:metric_learning_evaluation}. This suggests that such models could potentially be employed to assess the quality of generated outputs, reducing the reliance on annotations. These findings align with recent research on using LLMs as judges for generated outputs, such as the more general PanelGPT~\citepalias{sun2023query} and~\citetalias{li-etal-2024-side}'s work focussed on argument mining. They have also explored the capabilities of large language models in evaluating generated texts. While these results are promising, further research is needed to fully understand the implications and potential applications of these advancements specifically for evaluating the quality of explanations from a human perspective.

The contribution from our study can also be viewed from the angle of learning human preferences in natural language processing tasks. Unlike approaches that focus on developing automated metrics and maximising their correlations with human judgements~\citepalias{sellam2020bleurt}, our work investigates the ability to directly learn from human judgements. This approach aligns with recent advancements in preference learning, such as the Direct Preference Optimization (DPO) method~\citepalias{rafailov2024direct}, which learns latent human preferences from pairwise rankings of two model outputs. Our method learns the preferences directly---albeit based on pre-defined quality criteria instead of pairwise comparison---which is similar to the use of a learned reward model for optimising LLMs to follow instructions~\citepalias{ouyang2022training}. These approaches represent a shift towards more human-centered evaluation and optimisation in generative model training. While we have not directly incorporated the human quality judgements into the training of generative models due to the low number of annotated samples, an intriguing direction for future research would be to scale up these annotation efforts and directly integrate them into the training process. 


The reliability of crowdsourcing for data collection and annotation has been a topic of debate, with studies highlighting its potential for inconsistency~\citepalias{roit2020controlled, beck2023quality}. To address these concerns, our approach relies on \emph{(a)} qualified workers which passed attention checks, reducing the probability of random annotations; \emph{(b)} filtering methods to remove low-quality annotations. These enhanced quality control measures ensure more reliable outcomes \citepalias{barai2024crowdsourcing}. Furthermore, we augment some of the annotations with generative AI tools such as ChatGPT, which have been optimised on human preference in general domains \citepalias{ouyang2022training}, thus reflecting human judgements. It has been shown in the literature and confirmed by our study, that this has the potential to improve 
the quality and consistency of crowd-sourced data~\citepalias{ding2023gpt}.

\defcitealias{michie2011behaviour}{Michie et al., 2011}
However, even after establishing rigorous controls for annotation quality, we caution against over-generalising our findings, and advise careful consideration of the aim of any given study and its corresponding experiment design. One important distinction is the difference between how individuals judge the veracity of information---which is what we investigate---and how they act upon it in real-world contexts---which we explicitly make no statement about. Our experimental evidence focusses on whether a participant finds a particular explanation convincing, but this does not necessarily translate into behavioural change outside of the controlled environment.
To make statements about the latter and distinguish between judgements and actions, more controlled human-centred studies are required~\citepalias{michie2011behaviour}.



A critical consideration in the development and deployment of AI systems, such as fact-checking systems, involves addressing ethical concerns related to fairness, bias and responsibility. Given the societal implications of automated fact-checking, especially in influencing public opinion and decision-making, it is important to recognise that models can perpetuate biases present in their training data. Our approach, which focusses on contextualising claims rather than making definitive judgements, is potentially less affected by these issues because it does not aim to filter or block content. Instead, we aim to provide context around the information, allowing for a more nuanced understanding. This reduces the likelihood of inadvertently censoring or dismissing perspectives by considering diverse viewpoints. However, the generation of explanations themselves might be biased and thus risk reinforcing stereotypes or majority opinions. One approach to mitigate this, is to rely on multiple trusted data sources, i.e., FullFact, FactCheck and the BBC in our study. While effective, these strategies cannot fully prevent issues like model hallucinations, as formal guarantees against such errors are lacking in deep neural networks~\citep{kalai2024calibrated}. Additionally, for metric learning tasks, we ensure a broad representation of viewpoints by employing annotators from varied backgrounds, which helps address fairness and reduce bias \citepalias{frenda2024perspectivist, fleisig-etal-2024-perspectivist}. However, even though we rely on a diverse set of annotators, in this study we extract majority votes and averages from their annotations as signal for our metric learning models. An exciting future research direction that is made possible with our collected data is to model the distribution of opinions, especially when annotators disagree, as this might highlight debatable issues or topics where no single correct answer exists.


}

\defcitealias{yao2023end}{Yao et al. (2023)}

\section{Conclusions and Future Work}

In this paper, we present our work on generating human-accessible explanations as well as a human-centered approach for automatically evaluating the generated explanations. To facilitate the development and evaluation of our approaches, two novel datasets were developed: one for generating explanations within the context of fact-checking, and the other for the automatic evaluation of these explanations, using human annotations.

Based on our results, we revisit and answer the research questions presented in Section~\ref{sec:introduction}:

\begin{itemize}
  \item[\textbf{RQ1:}] How effectively can transformer-based models generate human-accessible explanations?

  As shown by the qualitative analysis in Table~\ref{tbl:generative_full_expl_dataset} and the quantitative results in Table~\ref{tbl:led_and_t5_rouge}, the transformer-based models are effective in generating an explanation within the context of fact checking, when presented with good evidence. However, some details are omitted at times. Conversely, when noisy evidence is supplied instead, the models' performance decreased significantly (Table~\ref{tbl:t5_base_results}), showing signs of input copying and self-contradictions. Furthermore, our empirical results show that there is a correlation between increasing dataset size and model performance, as evidenced in Table~\ref{tbl:t5_base_results}.

  \item[\textbf{RQ2:}] To what extent can the evaluation of fact-checking explanations be automated to align with human judgements across various qualitative dimensions?

  Based on the results presented in Figure~\ref{fig:results}, it can be seen that automating the evaluation of fact-checking explanations to align with human judgements across different dimensions is feasible, but challenging.
  Transformer-based models, particularly \texttt{DeBERTa-xxlarge}, when fine-tuned on our annotated dataset, show moderately strong correlations with human ratings primarily for objective dimensions such as \texttt{article contradiction}, \texttt{self-contradiction} and \texttt{hallucination}. However, automating the assessment of more subjective dimensions like \texttt{convincingness} and \texttt{overall quality} of explanations remains a challenge, with only marginal performance improvements over statistical baselines. This indicates that while automation can be effective in certain dimensions, achieving perfect agreement with human judgements across all qualitative dimensions is a difficult task, highlighting the need for ongoing research in the area of aligning model outputs with human standards.
\end{itemize}

The implications of our research are two fold. First, from a \textit{theoretical perspective}, our research advances the field of explainable artificial intelligence (XAI) by developing models capable of generating human-understandable explanations for fact-checking verdicts. This addresses a crucial gap in most fact-checking approaches which primarily focus on providing only a verdict, without an explanation. By integrating human-centered evaluation methods, our work emphasises the importance of the sought-after alignment with human assessments~\citep{schlegel2022towards} and how AI systems can be made more interpretable and accountable. Second, from a \textit{practical perspective}, our work enhances the reliability and transparency of AI-driven fact-checking systems by enabling them to provide understandable explanations for their decisions. This contributes to promoting users' trust and encourages critical engagement with the rationale behind the verdicts, addressing misinformation more effectively. Such advancements have the potential to improve the landscape of digital information verification. Additionally, the metric learning models we developed have the potential to improve the efficiency and effectiveness of AI systems in delivering reliable and trustworthy explanations for their decisions. By automating this evaluation, the model aids in significantly reducing the time and resources needed for manual assessment.

One limitation of our study is the reliance on text-only evidence for generating explanations, which overlooks the increasingly multi-modal nature of information online. In today's digital era, misinformation often spreads through various media, which usually include images, videos and audio, making it crucial for fact-checking systems to consider these modalities. Such systems could provide more comprehensive and nuanced explanations, hence, improving their effectiveness. Future work should explore integrating multi-modal inputs to better reflect real-world information, similar to the work presented by \citetalias{yao2023end}.
Additionally, while our metric learning models demonstrate potential, their agreement with human judgement varies across different quality dimensions. This underscores the need for further refinement of these models, particularly in capturing the nuances of human judgements.

\section*{Acknowledgements}
The authors would like to acknowledge the use
of the Computational Shared Facility at The University of Manchester and the use of Imperial College Research Computing Service (DOI: \url{http://doi.org/10.14469/hpc/2232}). This work was partially funded by the European Union’s Horizon 2020 research and innovation action programme, via the AI4Media Open Call \#1 issued and executed under the AI4Media project (Grant Agreement no. 951911) and is supported by the National Research Foundation, Prime Minister's Office, Singapore under its Campus for Research Excellence and Technological Enterprise (CREATE) programme.

\bibliographystyle{cas-model2-names}

\bibliography{cas-refs}

\appendix
\section{Google FactCheck API data formats}\label{apd:appendix1}
\begin{figure}[ht!]
        \centering
        \includegraphics[width=0.9\linewidth]{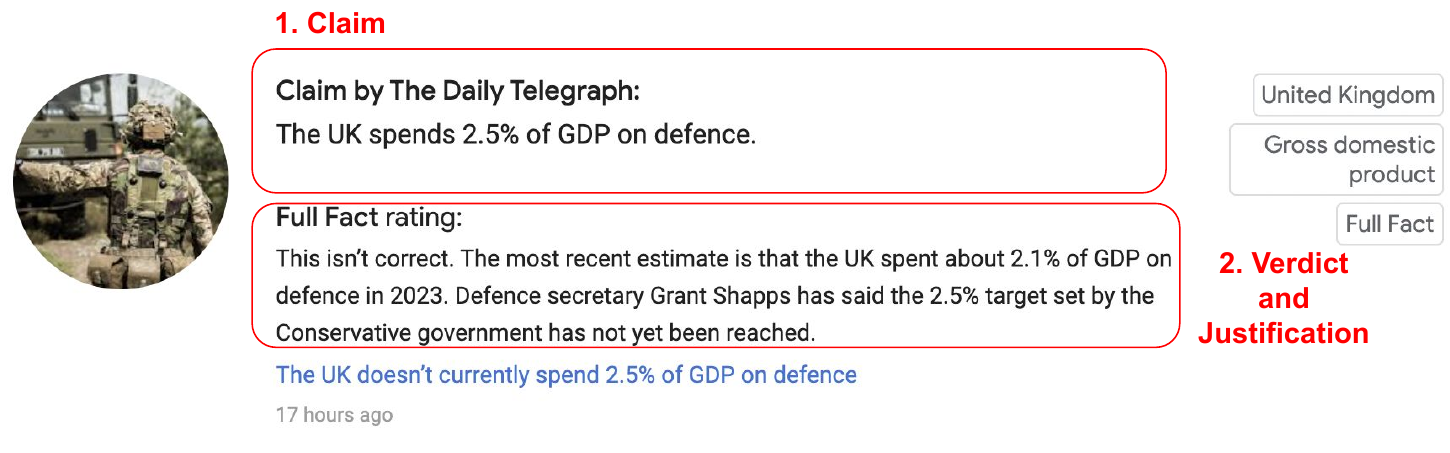}
        \caption{Data returned by the FactCheck API for \texttt{fullfact}. The data for \texttt{bbc} follows the same format.}
        \label{fig:sub1}
\end{figure}

\begin{figure}[ht!]
    \centering
    \includegraphics[width=1.0\linewidth]{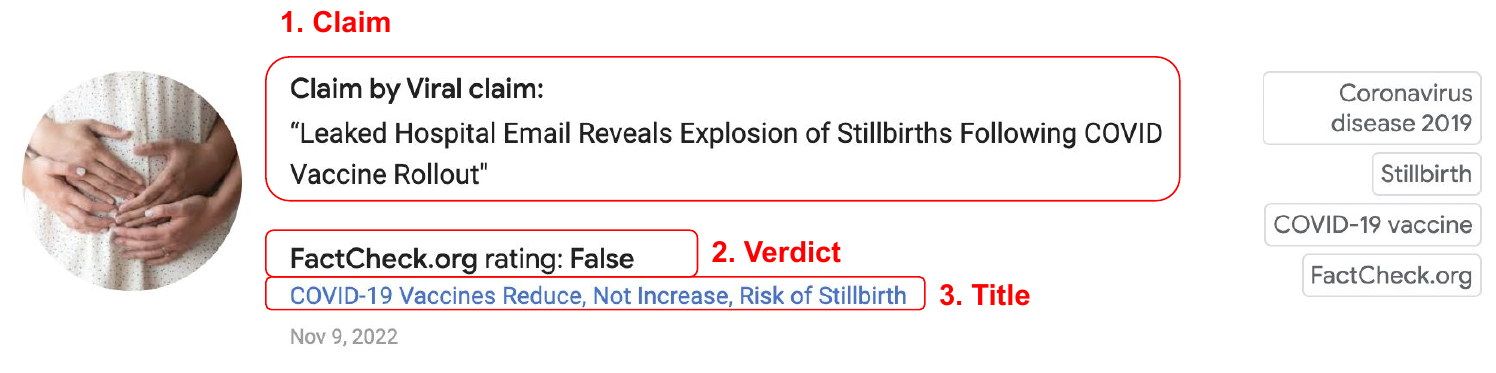}
    \caption{Data returned by the FactCheck API for \texttt{factcheck}. The title serves as explanation for the given claim.}
    \label{fig:sub2}
\end{figure}

See Figures~\ref{fig:sub1} and \ref{fig:sub2}.


\section{Mapping from textual verdicts to nominal categories}\label{apd:appendix_mapping} 
\begin{figure}[ht!]
    \centering\includegraphics[width=1.0\linewidth]{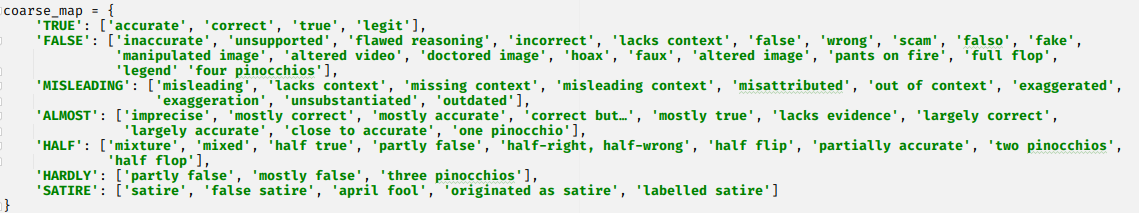}%
    \caption{Mapping from textual verdicts to nominal categories.}
    \label{fig:mapping}
\end{figure}

See Figure~\ref{fig:mapping}.
\section{Dimensions for the annotation of explanation quality}\label{apd:appendix_annotations_dimensions}
\begin{figure}[ht!]
    \centering    \includegraphics[width=0.5\linewidth]{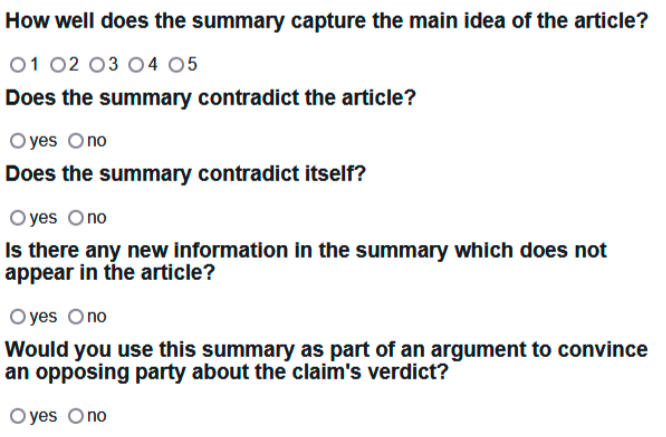}
    \caption{Specific questions in the annotation interface used to evaluate the explanation quality.}
    \label{fig:annotations_dimensions}

\end{figure}

See Figure~\ref{fig:annotations_dimensions}.
\newpage
\section{Annotation Interface}\label{apd:appendix_annotation_interface}
\begin{figure}[h]
    \centering
    \includegraphics[width=0.8\linewidth]{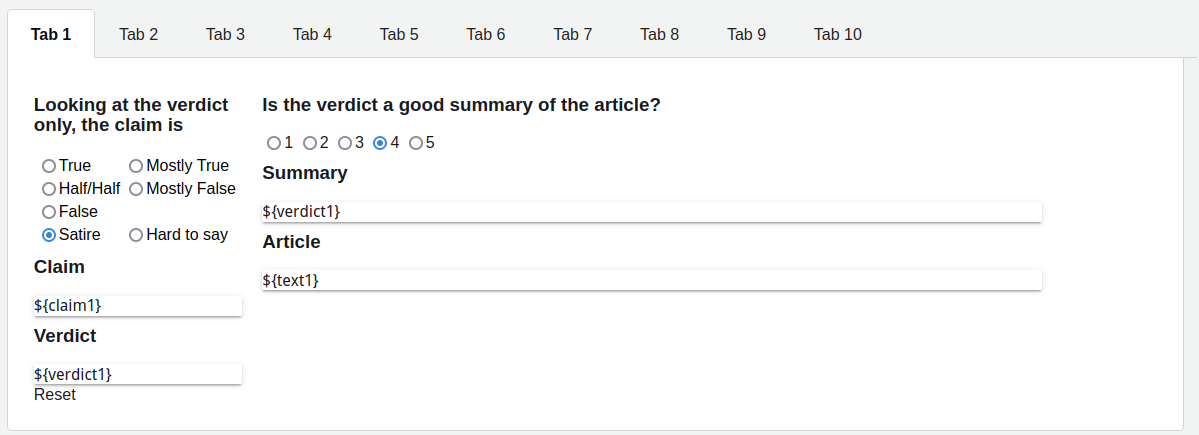}
    \caption{Interface for the qualification task. Once the verdict label is selected on the left, the right hand side of the interface appears to ask the crowdworker to judge the quality of the summary (i.e., to provide a verdict).}
    \label{fig:annotations_interface1}
\end{figure}

\begin{figure}[h]
    \centering
    \includegraphics[width=0.8\linewidth]{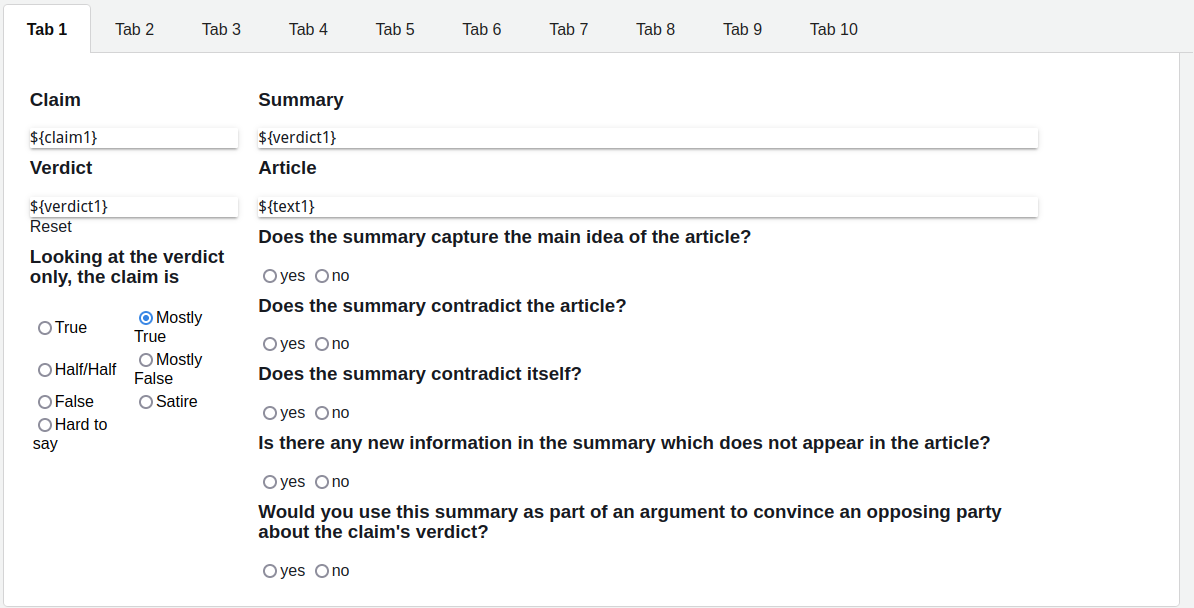}
    \caption{Interface for the annotation task. Once the verdict label is selected on the left, the right hand side of the interface appears to ask the crowdworker to judge the quality of the summary in the form of multiple binary questions.}
    \label{fig:annotations_interface2}
\end{figure}

See Figures~\ref{fig:annotations_interface1} and \ref{fig:annotations_interface2}.

\newpage

\section{Statistical Analysis for Metric Learning Models}\label{appendix:statistical_significance_metric_learning}

\begin{table}[h]
\caption{P-values for Classification Metric Learning Models}\label{tbl:metric_learning_pvalues}
\centering
\begin{tabular}{@{}ll@{}}
\toprule
\textbf{Comparison} & \textbf{P-value (MCC)} \\
\midrule
Convincingness: \texttt{DeBERTa-xxlarge} $<$ \texttt{DeBERTa-base} & 0.034 \\
Convincingness: \texttt{DeBERTa-xxlarge} $>$ Baseline & 0.032 \\
Convincingness: \texttt{DeBERTa-Base} $>$ Baseline & 0.004 \\
Article Contradiction: \texttt{DeBERTa-xxlarge} $>$ \texttt{DeBERTa-base} & 2.9e-5 \\
Article Contradiction: \texttt{DeBERTa-xxlarge} $>$ Baseline & 2.3e-6 \\
Article Contradiction: \texttt{DeBERTa-Base} $>$ Baseline & 5.9e-6 \\
Self-contradiction: \texttt{DeBERTa-xxlarge} $>$ \texttt{DeBERTa-base} & 0.02 \\
Self-contradiction: \texttt{DeBERTa-xxlarge} $>$ Baseline & 1.2e-6 \\
Self-contradiction: \texttt{DeBERTa-Base} $>$ Baseline & 0.005 \\
Hallucination: \texttt{DeBERTa-xxlarge} $>$ \texttt{DeBERTa-base} & 0.04 \\
Hallucination: \texttt{DeBERTa-xxlarge} $>$ Baseline & 3.1e-7 \\
Hallucination: \texttt{DeBERTa-Base} $>$ Baseline & 0.04 \\
\bottomrule
\end{tabular}
\end{table}

\begin{table}[h]
\caption{P-values for Regression Metric Learning Models. For MAE and MSE, $\cdot$ is $>$, for Spearman R, $\cdot$ is $<$.}\label{tbl:regression_pvalues}
\centering
\begin{tabular}{@{} llll @{} }
\toprule
\textbf{Comparison} & \textbf{P-value (MAE)} & \textbf{P-value (MSE)} & \textbf{Spearman R}\\
\midrule
\texttt{DeBERTa-Base} $\cdot$ \texttt{DeBERTa-xxlarge} & 0.01& 0.0004 & 0.42 \\
Baseline $\cdot$ \texttt{DeBERTa-Base} & 0.82 & 0.98 & 8.9e-6 \\
Baseline $\cdot$ \texttt{DeBERTa-xxlarge} & 0.16 & 0.60 & 0.0002 \\
\bottomrule
\end{tabular}
\end{table}
\newpage
\section{Hyper-parameters used for model training}\label{appendix:hyper-params}
Relevant hyper-parameters if not otherwise specified are further described in Tables~\ref{tbl:t5_hyperparams} and \ref{tbl:hyperparams}. For the sake of reproducibility, all code is accessible via \url{https://github.com/uomnlp/smaite-scripts}.
\begin{table}[h]
\caption{Hyperparameters for the Explanation Generation Model Training}\label{tbl:t5_hyperparams}
\centering
\begin{tabular}{@{}ll@{}}
\toprule
\textbf{Hyper-parameter} & \textbf{Value} \\
\midrule
Source Prefix           & summarize: \\
Max Input Length     & 1024 \\
Max Output Length     & 128 \\
Per Device Batch Size   & 8 \\
Learning Rate           & 5e-5 \\
Number of Epochs        & 3 \\
Optimiser               & AdamW \\
Learning Rate Scheduler & Warmup with Linear Decay \\
\bottomrule
\end{tabular}
\end{table}

\begin{table}[h]
\caption{Hyperparameters for the Metric Learning Model Training}\label{tbl:hyperparams}
\centering
\begin{tabular}{@{}ll@{}}
\toprule
\textbf{Hyper-parameter} & \textbf{Value} \\
\midrule
Inputs             & claim verdict text \\
Max Sequence Length     & 512 \\
Per Device Batch Size   & 4 \\
Learning Rate           & 3e-6 \\
Number of Epochs        & 4 \\
Warmup Steps            & 40 \\
Optimiser               & AdamW \\
Learning Rate Scheduler & Warmup with Linear Decay \\
\bottomrule
\end{tabular}
\end{table}

\end{document}